\documentclass[letterpaper]{article}
\usepackage{aaai25}
\usepackage{times}
\usepackage{helvet}
\usepackage{courier}
\usepackage[hyphens]{url}
\usepackage{graphicx}
\usepackage{natbib}
\usepackage{caption}
\usepackage{algorithm}
\usepackage{algorithmic}
\usepackage{bibentry}
\usepackage{newfloat}
\usepackage{listings}

\usepackage{booktabs}
\usepackage{pifont}
\usepackage{multirow}
\usepackage{amsmath}

\floatstyle{ruled}
\floatname{listing}{Listing}
\newfloat{listing}{tb}{lst}{}
\DeclareCaptionStyle{ruled}{
    labelfont=normalfont, 
    labelsep=colon, 
    strut=off
}
\title{ChatTime: A Unified Multimodal Time Series Foundation Model Bridging Numerical and Textual Data}

\author{
    Chengsen Wang\textsuperscript{\rm 1}\thanks{Equal contribution.} \ \ \ \ 
    Qi Qi\textsuperscript{\rm 1}\footnotemark[1] \ \ \ \ 
    Jingyu Wang\textsuperscript{\rm 1 \rm 2}\thanks{Corresponding author.} \\
    Haifeng Sun\textsuperscript{\rm 1} \ \ \ \ 
    Zirui Zhuang\textsuperscript{\rm 1} \ \ \ \ 
    Jinming Wu\textsuperscript{\rm 1} \ \ \ \ 
    Lei Zhang\textsuperscript{\rm 3} \ \ \ \ 
    Jianxin Liao\textsuperscript{\rm 1} \ \ \ \ 
}
\affiliations{
    \textsuperscript{\rm 1}Beijing University of Posts and Telecommunications, Beijing, China \\
    \textsuperscript{\rm 2}Pengcheng Laboratory, Shenzhen, China \\
    \textsuperscript{\rm 3}China Unicom Network Communications Corporation Limited, Beijing, China \\

    \texttt{\{cswang, qiqi8266, wangjingyu\}@bupt.edu.cn} \\
}

\begin{document}

\maketitle

\begin{abstract}

    Human experts typically integrate numerical and textual multimodal information to analyze time series. However, most traditional deep learning predictors rely solely on unimodal numerical data, using a fixed-length window for training and prediction on a single dataset, and cannot adapt to different scenarios. The powered pre-trained large language model has introduced new opportunities for time series analysis. Yet, existing methods are either inefficient in training, incapable of handling textual information, or lack zero-shot forecasting capability. In this paper, we innovatively model time series as a foreign language and construct ChatTime, a unified framework for time series and text processing. As an out-of-the-box multimodal time series foundation model, ChatTime provides zero-shot forecasting capability and supports bimodal input/output for both time series and text. We design a series of experiments to verify the superior performance of ChatTime across multiple tasks and scenarios, and create four multimodal datasets to address data gaps. The experimental results demonstrate the potential and utility of ChatTime. Code is available at https://github.com/ForestsKing/ChatTime.

\end{abstract}

\section{Introduction}
\label{section:Introduction}

Time series data is common in various fields, and its accurate forecasts are vital for decision support in industries such as finance \cite{HeSS23}, transportation \cite{HeZBYN22}, energy \cite{PintoPVS21}, healthcare \cite{PuriKVL22}, and climate \cite{DuLYH21}. Human experts frequently integrate multimodal information for time series forecasting. For instance, economists combine historical financial series with policy reports to predict future market trends. Due to their remarkable performance, deep learning predictors \cite{PatchTST, MSGNet} have become the mainstream method in recent years. However, most current deep paradigms train and predict on a single dataset based on fixed history and prediction windows, lacking adaptability to different scenarios or datasets. Additionally, most existing methods utilize only unimodal numerical data. Recent studies have demonstrated that simple linear models \cite{DLinear, RLinear} often rival the performance of state-of-the-art (SOTA) complex models, indicating that current unimodal approaches may be nearing a saturation point.

\begin{table}[t]
    \centering
    \setlength{\tabcolsep}{2mm}
    {\small
       \begin{tabular}{@{}c|cccc@{}}
            \toprule
            Method   & \begin{tabular}[c]{@{}c@{}}Zero-Shot \\ Forecast\end{tabular} & \begin{tabular}[c]{@{}c@{}}Missing \\ Support\end{tabular} & \begin{tabular}[c]{@{}c@{}}Training \\ Token\end{tabular} & \begin{tabular}[c]{@{}c@{}}Trainable \\ Parameter\end{tabular} \\ \midrule
            TimesFM  & \ding{51}                                                             & \ding{55}                                                          & 3T                                                           & 200M                                                            \\
            Moirai   & \ding{51}                                                             & \ding{51}                                                          & 150B                                                         & 300M                                                            \\
            TimeGPT  & \ding{51}                                                             & \ding{51}                                                          & 100B                                                         & Unknown                                                         \\
            MOMENT   & \ding{55}                                                             & \ding{51}                                                          & 100B                                                         & 300M                                                            \\
            Timer    & \ding{55}                                                             & \ding{55}                                                          & 50B                                                          & 50M                                                             \\
            Chronos  & \ding{51}                                                             & \ding{51}                                                          & 25B                                                          & 700M                                                            \\
            ChatTime & \ding{51}                                                             & \ding{51}                                                          & 1B                                                           & 350M                                                            \\ \bottomrule
        \end{tabular}
    }
    \caption{The comparison between pre-trained time series foundation models.}
    \label{tab01}
\end{table}

Meanwhile, the rapid advancement of pre-trained large language models (LLM) has garnered significant attention \cite{LLaMA, LLaMA2}. Through autoregressive pre-training on vast amounts of text, these robust tools are capable of performing a wide array of tasks in a zero-shot learning paradigm. This has spurred interest in incorporating LLMs into time series analysis. Some works \cite{Chronos, TimesFM} have utilized extensive time series data to construct time series foundational models, which can handle the forecasting task across any scenario with a single model. However, the training-from-scratch strategy renders them highly inefficient and forfeits the ability to process textual information. Other research \cite{TimeLLM, TGForecaster} has attempted to integrate the weights of pre-trained LLMs into a new time series forecasting framework. They fine-tune additional input and output layers to consider both time series and textual information. Nevertheless, these additional layers are incapable of zero-shot learning and require re-fine-tuning for each dataset. Furthermore, the inability to output text hindered the aforementioned paradigms in addressing scenarios such as time series question answering and summarization. This motivates the question: \textit{Is it possible to construct a multimodal time series foundation model that allows for zero-shot inference and supports both time series and textual bimodal inputs and outputs?}

Linguistic models for predicting the next word and time series models for predicting the next value fundamentally model the sequential structure of historical data to predict future patterns. At the core of both is an $n$-order Markov process \cite{GPT4TS}. In this work, we innovatively conceptualize time series as a foreign language and construct ChatTime, an out-of-the-box multimodal time series foundation model, as a framework for the unified processing of time series and text. ChatTime converts continuous unbounded time series into a finite set of discrete values through normalization and discretization, and then characterizes them as foreign language words by adding mark characters. We employ continuous pre-training and instruction fine-tuning for the pre-trained LLM using the same methodology as vocabulary expansion \cite{SambaLingo, EEVE}, eliminating the need to train from scratch or alter the model architecture. Compared to other foundation models, as shown in Table \ref{tab01}, we not only significantly reduce the training cost but also gain an additional inference capability to process textual information. This simple yet effective approach addresses a wide range of time series problems at minimal cost, paving the way for further leveraging the findings of LLMs and multimodal communities in the future.

To comprehensively evaluate the performance of ChatTime, we design a series of experiments including three main tasks: zero-shot time series forecasting (ZSTSF), context-guided time series forecasting (CGTSF), and time series question answering (TSQA). These tasks examine the modal translation capabilities of the foundational model for time series to time series, text to time series, and time series to text, respectively. Alongside the text-to-text inference capability of the pre-trained LLM itself \cite{GPT4}, ChatTime achieves seamless input and output of both time series and text modalities. The zero-shot time series forecasting task is evaluated on eight real-world benchmark datasets across four domains, which are commonly used \cite{Autoformer} for long-term time series forecasting. For the multimodal context-guided time series forecasting task, we collect time series records from three different scenarios, adding and aligning background, weather, and date information without any leakage of future information. Regarding the multimodal time series question answering task, we synthesize a variable-length question answering dataset covering four typical time series features \cite{TSFUBenchmark}. The experimental results confirm the superior performance of ChatTime in multiple tasks and scenarios, highlighting its potential as a multimodal time series foundation model.

In general, the contributions of our paper are summarised as follows:

\begin{itemize}

    \item We construct ChatTime, a multimodal time series foundation model, by conceptualizing time series as a foreign language. It allows for zero-shot inference and supports both time series and textual bimodal inputs and outputs.
    
    \item We establish three context-guided time series forecasting datasets and a time series question answering dataset to fill gaps in related multimodal domains, offering valuable resources for future research.
    
    \item We demonstrate the considerable advantages of ChatTime across multiple time series tasks through comprehensive experiments, offering innovative perspectives and solutions for time series analysis.
    
\end{itemize}

\section{Related Work}
\label{section:Related-Work}

\subsection{Long-Term Time Series Forecasting}
\label{subsection:Long-Term-Time-Series-Forecasting}

As a significant real-world challenge, time series forecasting has garnered considerable attention. Initially, ARIMA \cite{ARIMA} performs forecasts in a moving average manner. However, the complex real world often renders such statistical methods challenging to adapt. With the development of deep learning, neural network-based methods have become increasingly important. Recurrent neural networks \cite{LSTM, DeepAR} dynamically capture temporal dependencies within a sequential structure. Unfortunately, this architecture suffers from gradient vanishing/exploding and information forgetting. To further improve prediction performance, convolutional networks \cite{MICN, TimesNet} and self-attention mechanisms \cite{Informer, iTransformer} have been introduced to capture long-range dependencies. Despite achieving impressive performance, most current deep paradigms lack adaptability to different scenarios and utilize only unimodal numerical data.

\subsection{LLM-Based Time Series Analysis}
\label{subsection:LLM-Based-Time-Series-Analysis}

The rise of pre-trained LLMs has introduced new opportunities for time series analysis. Based on the dependence on pre-training weights, these works can be broadly categorized into the following three paradigms.

The first category of work relies entirely on pre-trained weights. They \cite{LLMTIME} employ LLMs directly for time series forecasting via prompts. Due to the lack of understanding \cite{TSFUBenchmark} about time series features, their prediction accuracy is typically too low \cite{TSandLanguage}. These methods also have low token utilization due to the bit-by-bit tokenization. Instruction fine-tuning has improved accuracy in some cases \cite{TPLLM}, but these improvements do not address high inference costs.

The second category of work integrates pre-training weights into new frameworks. Additional neural layers will be fine-tuned to adapt for the time series. Some studies \cite{GPT4TS, TimeLLM} use pre-trained weights as the backbone and incorporate extra input and output layers, significantly enhancing prediction performance. Others \cite{TGForecaster, GPT4MTS} utilize pre-trained weights as an embedding module to enable the reception of context. However, most of them cannot perform zero-shot inference.

The third category of work uses the architecture of pre-trained LLMs but does not utilize the weights. They \cite{TimeGPT, Chronos, TimesFM, Moirai, MOMENT, Timer} employ vast amounts of time series data to construct new foundation models. While yielding promising results, training from scratch is highly inefficient, and most of these models support only unimodal numerical data.

Some studies have explored the multimodal time series pre-training within limited domains and tasks \cite{MedicalFM, METS}. Plotting time series into charts \cite{ChartAssisstant, ChartInstruct} is also viable. However, they do not support fine-grained time series forecasting, the most crucial task of time series analysis.

\begin{figure*}[t]
    \centering
    \resizebox{0.975\linewidth}{!}
    {
        \includegraphics{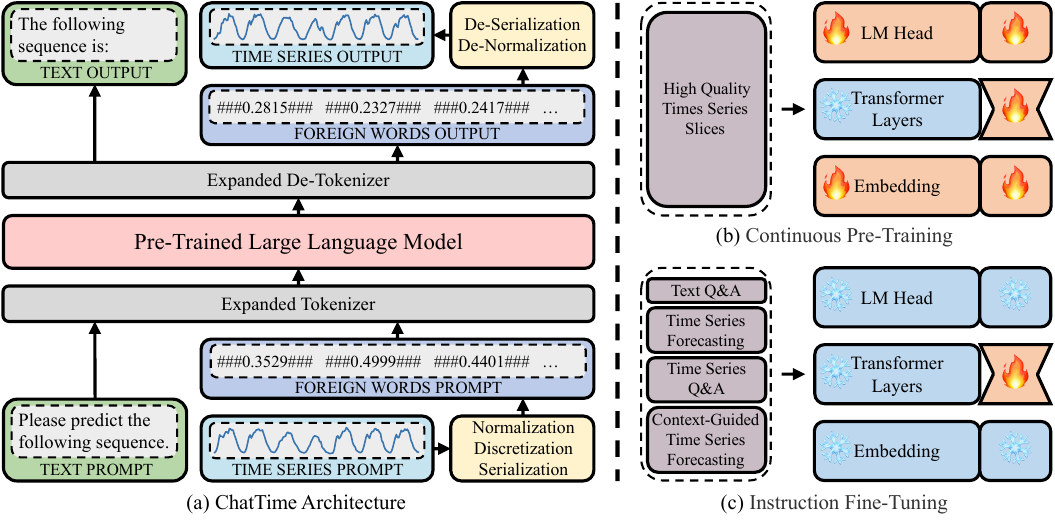}
    }
    \caption{The overview of ChatTime. (a) illustrates the overall architecture, introducing the yellow plug-ins that enable the intertranslation of time-series real values and foreign language. The vocabulary of the grey tokenizer is also extended to accommodate the time series language. We further pre-train (b) and fine-tune (c) existing LLMs using the same methodology as vocabulary expansion, eliminating the need to train from scratch or alter the model architecture.}
    \label{fig01}
\end{figure*}

\section{Methodology}
\label{section:Methodology}

\subsection{Overview}
\label{subsection:Overview}

As illustrated in Figure \ref{fig01}(a), ChatTime initially encodes time series into a foreign language through normalization, discretization, and the incorporation of mark characters. The expanded tokenizer then transforms text and foreign words into token indexes. After processing by the LLM, the de-tokenizer translates the token indexes back into text and foreign words. Finally, the foreign words are re-decoded into time series by removing mark characters and applying inverse normalization. As depicted in Figures \ref{fig01}(b) and \ref{fig01}(c), the training process is divided into two phases: continuous pre-training and instruction fine-tuning. Both phases utilize 4-bit quantized models with LoRA \cite{LoRA}.

\subsection{Model Architecture}
\label{subsection:Model-Architecture}

By conceptualizing it as a foreign language, ChatTime enables pre-trained LLMs to process time series through vocabulary expansion. As illustrated in Figure \ref{fig01}(a), ChatTime implements two critical modifications: first, it introduces a yellow plug-in that supports the interconversion between real values of time series and foreign language; second, it extends the vocabulary of grey tokenizer to accommodate time series language.

Unlike natural language derived from a finite dictionary, time series are typically real-valued data within unbounded continuous domains. Consider the time series $\mathbf{x}_{1:C+H}=\left\{x_1, \ldots, x_{1:C+H}\right\}$, where the initial $C$ time steps constitute the history series, and the subsequent $H$ time steps form the prediction series. ChatTime employs min-max scaling to map unbounded real values into a bounded range of -1 to 1. Given that the prediction series is unknown during the actual inference process, we scale based solely on the history series. Acknowledging that the prediction series may surpass the range of the history series, we scale the history series into the range of -0.5 to 0.5, reserving the remaining interval as a buffer for the prediction series. The scaling process is described as follows: \begin{equation}
    \begin{gathered}
        \begin{aligned}
            \tilde{\mathbf{x}}_{1:C+H} = \frac{\mathbf{x}_{1:C+H}-\operatorname{min}\left(\mathbf{x}_{1:C}\right)}{\operatorname{max}\left(\mathbf{x}_{1:C}\right)-\operatorname{min}\left(\mathbf{x}_{1:C}\right)}-0.5
        \end{aligned}
    \end{gathered}
\end{equation}

The scaled time series remain continuous real values that cannot be directly converted into a finite dictionary. We employ a binning technique to quantize these real values into discrete tokens. Specifically, we uniformly partition the interval from -1 to 1 into 10K bins. Each scaled real value is mapped to the corresponding bin, and the center value of the bin is used as the quantized lossy discrete value.

\begin{table}[t]
    \centering
    \setlength{\tabcolsep}{1mm}
    {\small
        \begin{tabular}{@{}l@{}}
            \toprule
            \textbf{Time Series}                                                                                                                                                                                                                              \\
            \textit{{[}0.2835, 0.2285, 0.1587, 0.4001{]}}                                                                                                                                                                                                     \\ \midrule
            \textbf{GPT (34 tokens)}                                                                                                                                                                                                                          \\
            \textit{"2 8 3 5 , 2 2 8 5 , 1 5 8 7 , 4 0 0 1"}                                                                                                                                                                                                  \\
            \begin{tabular}[c]{@{}l@{}}{[}'2', '\_', '8', '\_', '3', '\_', '5', '\_,', '\_', '2', '\_', '2', \\ '\_', '8', '\_', '5', '\_,', '\_', '1', '\_', '5', '\_', '8', \\ '\_', '7', '\_,', '\_', '4', '\_', '0', '\_', '0', '\_', '1'{]}\end{tabular} \\ \midrule
            \textbf{LLaMA (22 tokens)}                                                                                                                                                                                                                        \\
            \textit{"2835, 2285, 1587, 4001"}                                                                                                                                                                                                                 \\
            \begin{tabular}[c]{@{}l@{}}{[}'2', '8', '3', '5', ',', '\_', '2', '2', '8', '5', ',', \\ '\_', '1', '5', '8', '7', ',', '\_', '4', '0', '0', '1'{]}\end{tabular}                                                                                  \\ \midrule
            \textbf{ChatTime (7 tokens)}                                                                                                                                                                                                                      \\
            \textit{"\#\#\#0.2835\#\#\# \#\#\#0.2285\#\#\# \#\#\#0.1587\#\#\# \#\#\#0.4001\#\#\#"}                                                                                                                                                            \\
            \begin{tabular}[c]{@{}l@{}}{[}'\#\#\#0.2835\#\#\#', '\_', '\#\#\#0.2285\#\#\#', '\_', \\ '\#\#\#0.1587\#\#\#', '\_', '\#\#\#0.4001\#\#\#'{]}\end{tabular}                                                                                         \\ \bottomrule
        \end{tabular}
    }
    \caption{The comparison of token consumption between LLMTIME and ChatTime.}
    \label{tab02}
\end{table}

Next, we fix the precision of the discretized time series to 4 like LLMTIME \cite{LLMTIME}. As illustrated in Table \ref{tab02}, LLMTIME presents two methods for GPT and LLaMA tokenizing time series bit-by-bit. However, this method consumes a substantial number of tokens, leading to large computational costs. To address this issue, we introduce the mark characters \textit{"\#\#\#"} at the beginning and end of the discretized time series to form foreign language words. By extending the vocabulary of the tokenizer, only one token is needed for each value, regardless of its precision. Moreover, not only do we add the foreign words derived from the center of the 10K bins into the vocabulary, but also include an additional \textit{"\#\#\#Nan\#\#\#"} to manage missing values.

\subsection{Continuous Pre-Training}
\label{subsection:Continuous-Pre-Training}

Continuous pre-training is frequently employed to enhance the comprehension of LLMs in specialized domains. Grasping the fundamental principles of time series is essential for executing downstream tasks. As depicted in Figure \ref{fig01}(b), during the continuous pre-training stage, 1M high quality time series slices are used to pre-train LLaMA-2-7B-Base \cite{LLaMA2}, resulting in ChatTime-1-7B-Base. We employ autoregressive forecasting on extensive time series data as a pre-training task. As the vocabulary of the tokenizer is expanded, the embedding layer and output header also require training alongside the Transformer layer.

The data for continuous pre-training is sourced from two extensive open-source time series repositories, Monash \cite{Monash} and TFB \cite{TFB}, encompassing approximately 100 sub-datasets. Notably, the 11 sub-datasets for evaluating ZSTSF and CGTSF tasks in Section \ref{subsection:Zero-Shot-Time-Series-Forecasting-Experiment} and \ref{subsection:Context-Guided-Time-Series-Forecasting-Experiment} have been excluded to prevent information leakage. The autoregressive forecasting strategy enables ChatTime to support history and prediction windows of any size. We apply sliding slices to the original time series using five distinct window and step sizes, as illustrated in Table \ref{tab03}. We prioritize slicing the original time series into larger segments. Given the numerous repeating patterns and the limited computational resources, we perform K-means \cite{Kmeans} on 10M original time series slices. We categorize them into 1M and 25K groups, randomly selecting one sample from each group to serve as a representative. Consequently, we create a high-quality dataset for continuous pre-training (1M) and instruction fine-tuning (25K).

\begin{table}[t]
    \centering
    \setlength{\tabcolsep}{1mm}
    {\small
        \begin{tabular}{@{}ccccc@{}}
            \toprule
            Window Size & History Length & Prediction Length & Sliding Step \\ \midrule
            576         & 512            & 64                & 32           \\
            288         & 256            & 32                & 16           \\
            144         & 128            & 16                & 8            \\
            72          & 64             & 8                 & 4            \\
            36          & 32             & 4                 & 2            \\ \bottomrule
        \end{tabular}
    }
    \caption{The setting of sliding windows when constructing continuous pre-training dataset.}
    \label{tab03}
\end{table}

\subsection{Instruction Fine-Tuning}
\label{subsection:Instruction-Fine-Tuning}

As shown in Figure \ref{fig01}(c), during the instruction fine-tuning phase, four task datasets are used to fine-tune ChatTime-1-7B-Base, yielding the final ChatTime-1-7B-Chat. 25K samples are extracted for each task, totaling 100K instances of fine-tuned data. We only fine-tune the Transformer layer during this phase.

We introduce the text question answering task to retain the textual inference capabilities of the LLMs. We randomly select 25K samples from the widely used Alpaca \cite{Alpaca} dataset for this task. For the unimodal time series forecasting task, we utilize 25K high quality time series slices from Section \ref{subsection:Continuous-Pre-Training}. Moreover, context-guided forecasting and time series question answering tasks involve the interconversion of time series and text modalities, where related datasets are lacking. Therefore, we collect three CGTSF datasets and synthesize a TSQA dataset to address this gap and offer a valuable resource for future research.

The context-guided forecasting task is supported by three multimodal datasets: Melbourne Solar Power Generation (MSPG), London Electricity Usage (LEU), and Paris Traffic Flow (PTF). Only background, weather (forecast from Open-Meteo \cite{OpenMeteo}), and date are included as textual auxiliary information to prevent future information leakage. Detailed dataset information is provided in Appendix \ref{subsection:Context-Guided-Time-Series-Forecasting-Dataset}. To avoid information leakage during the evaluation phase in Section \ref{subsection:Context-Guided-Time-Series-Forecasting-Experiment}, each dataset is chronologically split into training, validation, and test sets with a ratio of 6:2:2. A sample of 25K data points is randomly selected from the training sets of these three datasets.

For the time series question answering task, we employ the KernelSynth \cite{Chronos} to generate a variable-length multimodal question answering dataset based on four generic typical time series features \cite{TSFUBenchmark}. Detailed dataset information is provided in Appendix \ref{subsection:Time-Series-Question-Answering-Dataset}. We randomly select 25K data entries from this dataset for instruction fine-tuning. By aligning time series features with textual representations, this task can also improve the performance of ChatTime in context-guided forecasting.

\begin{table*}[t]
    \centering
    \setlength{\tabcolsep}{2mm}
    {\small
       \begin{tabular}{@{}ccc|cccc|ccccc@{}}
            \toprule
            \multicolumn{1}{c|}{\multirow{2}{*}{Dataset}}     & \multirow{2}{*}{Hist} & \multirow{2}{*}{Pred} & \multicolumn{4}{c|}{Full-Shot Forecast}                               & \multicolumn{5}{c}{Zero-Shot Forecast}                                   \\ \cmidrule(l){4-12} 
            \multicolumn{1}{c|}{}                             &                       &                       & DLinear         & iTransformer    & GPT4TS          & TimeLLM         & TimeGPT      & Moirai       & TimesFM      & Chronos      & ChatTime     \\ \midrule
            \multicolumn{1}{c|}{\multirow{4}{*}{\rotatebox{90}{ETTh1}}}       & 48    & 24                    & 0.1462          & 0.1650          & \textbf{0.1389}          & 0.1467          & \textbf{0.1604}       & 0.1694       & 0.2021       & 0.1634       & 0.1698       \\
            \multicolumn{1}{c|}{}                             & 72                    & 24                    & \textbf{0.1358}          & 0.1852          & 0.1469          & 0.1439          & 0.1603       & 0.1796       & 0.1599       & \textbf{0.1372}       & 0.1403       \\
            \multicolumn{1}{c|}{}                             & 96                    & 24                    & \textbf{0.1398}          & 0.1964          & 0.1447          & 0.1473          & 0.1577       & 0.1433       & 0.1454       & \textbf{0.1374}       & 0.1374       \\
            \multicolumn{1}{c|}{}                             & 120                   & 24                    & \textbf{0.1371}          & 0.1971          & 0.1414          & 0.1513          & 0.1594       & 0.1492       & 0.1502       & \textbf{0.1348}       & 0.1431       \\ \midrule
            \multicolumn{1}{c|}{\multirow{4}{*}{\rotatebox{90}{ETTh2}}}       & 48    & 24                    & \textbf{0.2724}          & 0.2937          & 0.2742          & 0.2758          & \textbf{0.2874}       & 0.2963       & 0.3360       & 0.3128       & 0.2906       \\
            \multicolumn{1}{c|}{}                             & 72                    & 24                    & 0.2756          & 0.3118          & \textbf{0.2717}          & 0.2972          & 0.2888       & 0.3109       & \textbf{0.2880}       & 0.3045       & 0.3092       \\
            \multicolumn{1}{c|}{}                             & 96                    & 24                    & \textbf{0.2831}          & 0.3417          & 0.2900          & 0.2864          & \textbf{0.2902}       & 0.3139       & 0.3144       & 0.3158       & 0.2917       \\
            \multicolumn{1}{c|}{}                             & 120                   & 24                    & 0.2863          & 0.3299          & \textbf{0.2854}          & 0.3175          & 0.3026       & \textbf{0.2905}       & 0.3311       & 0.3150       & 0.3124       \\ \midrule
            \multicolumn{1}{c|}{\multirow{4}{*}{\rotatebox{90}{ETTm1}}}       & 192   & 96                    & 0.1479          & 0.1608          & \textbf{0.1384}          & 0.1503          & 0.1921       & 0.1608       & 0.1719       & 0.1604       & \textbf{0.1442}       \\
            \multicolumn{1}{c|}{}                             & 288                   & 96                    & 0.1400          & 0.1813          & \textbf{0.1345}          & 0.1425          & 0.1715       & 0.1848       & 0.1650       & \textbf{0.1452}       & 0.1587       \\
            \multicolumn{1}{c|}{}                             & 384                   & 96                    & \textbf{0.1428}          & 0.1680          & 0.1518          & 0.1452          & 0.1616       & 0.1619       & 0.1584       & 0.1463       & \textbf{0.1393}       \\
            \multicolumn{1}{c|}{}                             & 480                   & 96                    & \textbf{0.1406}          & 0.2001          & 0.1472          & 0.1527          & 0.1570       & 0.1703       & 0.1582       & \textbf{0.1401}       & 0.1802       \\ \midrule
            \multicolumn{1}{c|}{\multirow{4}{*}{\rotatebox{90}{ETTm2}}}       & 192   & 96                    & 0.2793          & 0.3397          & \textbf{0.2792}          & 0.2918          & 0.4294       & 0.4206       & 0.3405       & 0.3759       & \textbf{0.3135}       \\
            \multicolumn{1}{c|}{}                             & 288                   & 96                    & \textbf{0.2881}          & 0.3623          & 0.2918          & 0.2904          & 0.3625       & 0.3882       & \textbf{0.3277}       & 0.3472       & 0.3340       \\
            \multicolumn{1}{c|}{}                             & 384                   & 96                    & 0.2947          & \textbf{0.2880}          & 0.3089          & 0.3003          & \textbf{0.3389}       & 0.3742       & 0.3562       & 0.3589       & 0.3434       \\
            \multicolumn{1}{c|}{}                             & 480                   & 96                    & 0.3014          & 0.3725          & \textbf{0.2945}          & 0.3054          & \textbf{0.3242}       & 0.3597       & 0.3679       & 0.3353       & 0.4213       \\ \midrule
            \multicolumn{1}{c|}{\multirow{4}{*}{\rotatebox{90}{Electric}}}    & 48    & 24                    & 0.5719          & 0.5951          & \textbf{0.5008}          & 0.5733          & \textbf{0.5276}       & 0.6617       & 0.6005       & 0.6098       & 0.6083       \\
            \multicolumn{1}{c|}{}                             & 72                    & 24                    & 0.5486          & 0.5619          & \textbf{0.4896}          & 0.4989          & \textbf{0.4953}       & 0.6018       & 0.5454       & 0.5914       & 0.6238       \\
            \multicolumn{1}{c|}{}                             & 96                    & 24                    & 0.5536          & 0.5290          & \textbf{0.4432}          & 0.4816          & 0.4971       & 0.5260       & 0.5276       & 0.5139       & \textbf{0.4951}       \\
            \multicolumn{1}{c|}{}                             & 120                   & 24                    & 0.4714          & 0.5622          & \textbf{0.4540}          & 0.4848          & 0.5196       & 0.4963       & \textbf{0.4900}       & 0.5031       & 0.5101       \\ \midrule
            \multicolumn{1}{c|}{\multirow{4}{*}{\rotatebox{90}{Exchange}}}    & 14    & 7                     & 0.0543          & \textbf{0.0526}          & 0.0533          & 0.0531          & 0.0620       & 0.0784       & 0.0647       & 0.0555       & \textbf{0.0540}       \\
            \multicolumn{1}{c|}{}                             & 21                    & 7                     & 0.0571          & 0.0547          & \textbf{0.0505}          & 0.0505          & 0.0599       & 0.0812       & 0.0743       & 0.0635       & \textbf{0.0556}       \\
            \multicolumn{1}{c|}{}                             & 28                    & 7                     & 0.0595          & 0.0581          & \textbf{0.0508}          & 0.0511          & 0.0610       & 0.0844       & 0.0652       & 0.0595       & \textbf{0.0559}       \\
            \multicolumn{1}{c|}{}                             & 35                    & 7                     & 0.0615          & 0.0607          & \textbf{0.0493}          & 0.0524          & 0.0629       & 0.0677       & 0.0632       & 0.0598       & \textbf{0.0558}       \\ \midrule
            \multicolumn{1}{c|}{\multirow{4}{*}{\rotatebox{90}{Traffic}}}     & 48    & 24                    & 0.4662          & 0.5000          & 0.4557          & \textbf{0.4473}          & 0.4668       & 0.4887       & 0.4483       & 0.4718       & \textbf{0.4220}       \\
            \multicolumn{1}{c|}{}                             & 72                    & 24                    & 0.4475          & 0.4443          & \textbf{0.4116}          & 0.4252          & 0.4635       & 0.4581       & 0.4196       & \textbf{0.3725}       & 0.3873       \\
            \multicolumn{1}{c|}{}                             & 96                    & 24                    & 0.4438          & 0.4348          & 0.4190          & \textbf{0.4064}          & 0.4332       & 0.4082       & \textbf{0.3714}       & 0.3787       & 0.4074       \\
            \multicolumn{1}{c|}{}                             & 120                   & 24                    & 0.4190          & 0.4149          & \textbf{0.3416}          & 0.4279          & 0.4161       & \textbf{0.3539}       & 0.3542       & 0.3908       & 0.4125       \\ \midrule
            \multicolumn{1}{c|}{\multirow{4}{*}{\rotatebox{90}{Weather}}}     & 288   & 144                   & \textbf{0.0339}          & 0.0367          & 0.0364          & 0.0352          & 0.0331       & \textbf{0.0305}       & 0.0354       & 0.0343       & 0.0352       \\
            \multicolumn{1}{c|}{}                             & 432                   & 144                   & \textbf{0.0366}          & 0.0404          & 0.0401          & 0.0395          & 0.0321       & 0.0302       & \textbf{0.0298}       & 0.0346       & 0.0356       \\
            \multicolumn{1}{c|}{}                             & 576                   & 144                   & \textbf{0.0364}          & 0.0379          & 0.0399          & 0.0377          & 0.0328       & 0.0331       & 0.0321       & 0.0349       & \textbf{0.0284}       \\
            \multicolumn{1}{c|}{}                             & 720                   & 144                   & \textbf{0.0371}          & 0.0395          & 0.0392          & 0.0392          & \textbf{0.0323}       & 0.0353       & 0.0369       & 0.0335       & 0.0332       \\ \midrule
            \multicolumn{3}{c|}{Avg. MAE}                                                                     & 0.2409          & 0.2661          & \textbf{0.2286}          & 0.2390          & 0.2544       & 0.2659       & 0.2541       & \textbf{0.2512}       & 0.2515       \\ \midrule
            \multicolumn{3}{c|}{Avg. Rank}                                                                    & 3.7500          & 6.9688          & \textbf{3.0000}          & 3.9688          & 5.5625       & 6.5000       & 5.7500       & 4.8438       & \textbf{4.4688}       \\ \bottomrule
        \end{tabular}
    }
    \caption{The evaluation result in the traditional unimodal time series forecasting task. The lower values for all metrics represent the better performance. The best results among full-shot and zero-shot forecasting methods are highlighted in bold, respectively.}
    \label{tab04}
\end{table*}

\section{Experiment}
\label{section:Experiment}

\subsection{Implementation Setting}
\label{subsection:Implementation-Settings}

The training process of ChatTime is divided into continuous pre-training and instruction fine-tuning. Both phases utilize 4-bit quantized models with LoRA. In the LoRA, the rank and alpha are set to 8 and 16, respectively. The batch size is 8 with a gradient accumulation of 32, resulting in a global batch size of 256. The number of epochs for pre-training is set to 2, spanning 8K steps, with a visualization of the losses shown in Figure \ref{fig03}(a). The number of epochs for fine-tuning is set to 4, spanning 1.6K steps, with a visualization of the losses depicted in Figure \ref{fig03}(b). Owing to Unsloth \cite{Unsloth}, the entire train process can be executed on an Ubuntu server equipped with a single NVIDIA GeForce RTX 4090 graphics card. All source code, data, and weight will be made publicly accessible upon the publication of the paper.

\subsection{Zero-Shot Time Series Forecasting}
\label{subsection:Zero-Shot-Time-Series-Forecasting-Experiment}

For the regular unimodal time series forecasting task, we conduct experiments on eight datasets across four domains: Electric, Exchange, Traffic, and Weather, in addition to four ETT datasets. These datasets, widely used for benchmarking, are publicly available \cite{Autoformer}. Detailed information is provided in Appendix \ref{subsection:Zero-Shot-Time-Series-Forecasting-Dataset}. Notably, we have excluded these datasets during the training of ChatTime to prevent information leakage. Each dataset is chronologically divided into training, validation, and test sets with a ratio of 6:2:2. We determine a priori period of each dataset based on its collection granularity and use it as the prediction length. The history length is set to be \{2,3,4,5\} times the prediction length, ensuring that the history window of the zero-shot models contains at least two complete periods. We report the Mean Absolute Error (MAE) as the evaluation metric, where lower values mean better performance.

\begin{table*}[t]
    \centering
    \setlength{\tabcolsep}{2mm}
    {\small
       \begin{tabular}{@{}ccc|cccc|ccccc@{}}
            \toprule
            \multicolumn{1}{c|}{\multirow{2}{*}{Dataset}} & \multirow{2}{*}{Hist} & \multirow{2}{*}{Pred} & \multicolumn{4}{c|}{Dataset-Specific Forecast}              & \multicolumn{5}{c}{Dataset-Shared Forecast}                           \\ \cmidrule(l){4-12} 
            \multicolumn{1}{c|}{}                         &                       &                       & DLinear         & GPT4TS & TimeLLM & TGForecaster    & Moirai & TimesFM & Chronos         & ChatTime- & ChatTime        \\ \midrule
            \multicolumn{1}{c|}{\multirow{4}{*}{\rotatebox{90}{MSPG}}}     & 192   & 96                    & \textbf{0.7136}          & 0.7558 & 0.7697  & 0.7595          & 0.8108 & 0.8362  & 0.7427          & 0.7606    & \textbf{0.7346}          \\
            \multicolumn{1}{c|}{}                         & 288                   & 96                    & \textbf{0.7083}          & 0.7464 & 0.7959  & 0.7610          & 0.7849 & 0.7896  & 0.7408          & 0.7606    & \textbf{0.7353}          \\
            \multicolumn{1}{c|}{}                         & 384                   & 96                    & \textbf{0.7014}          & 0.7388 & 0.7672  & 0.7638          & 0.7749 & 0.7811  & 0.7352          & 0.7607    & \textbf{0.7330}          \\
            \multicolumn{1}{c|}{}                         & 480                   & 96                    & \textbf{0.7018}          & 0.7311 & 0.7632  & 0.7695          & 0.7664 & 0.7667  & 0.7344          & 0.7607    & \textbf{0.7292}          \\ \midrule
            \multicolumn{1}{c|}{\multirow{4}{*}{\rotatebox{90}{LEU}}}     & 96    & 48                    & 0.6676          & 0.6697 & 0.6531  & \textbf{0.6181}          & \textbf{0.6228} & 0.6670  & 0.6571          & 0.6496    & 0.6305          \\
            \multicolumn{1}{c|}{}                         & 144                   & 48                    & 0.6495          & 0.6567 & 0.6474  & \textbf{0.6355}          & \textbf{0.6085} & 0.6475  & 0.6597          & 0.6506    & 0.6231          \\
            \multicolumn{1}{c|}{}                         & 192                   & 48                    & 0.6407          & 0.6771 & \textbf{0.6329}  & 0.6458          & \textbf{0.6008} & 0.6490  & 0.6645          & 0.6407    & 0.6111          \\
            \multicolumn{1}{c|}{}                         & 240                   & 48                    & \textbf{0.6316}          & 0.6383 & 0.6356  & 0.6329          & \textbf{0.5968} & 0.6333  & 0.6631          & 0.6377    & 0.6085          \\ \midrule
            \multicolumn{1}{c|}{\multirow{4}{*}{\rotatebox{90}{PTF}}}     & 48    & 24                    & 0.5204          & 0.4373 & \textbf{0.4211}  & 0.4411          & 0.5981 & 0.4851  & \textbf{0.4813}          & 0.5155    & 0.4849          \\
            \multicolumn{1}{c|}{}                         & 72                    & 24                    & 0.5075          & 0.4253 & 0.4031  & \textbf{0.3943}          & 0.5776 & \textbf{0.4258}  & 0.4276          & 0.4436    & 0.4307          \\
            \multicolumn{1}{c|}{}                         & 96                    & 24                    & 0.4965          & 0.3921 & 0.4392  & \textbf{0.3653}          & 0.5179 & 0.4054  & 0.4336          & 0.4172    & \textbf{0.3920}          \\
            \multicolumn{1}{c|}{}                         & 120                   & 24                    & 0.4796          & 0.3713 & 0.3594  & \textbf{0.3594}          & 0.5245 & 0.3807  & 0.3902          & 0.3943    & \textbf{0.3480}          \\ \midrule
            \multicolumn{3}{c|}{Avg. MAE}                                                                 & 0.6182          & 0.6033 & 0.6073  & \textbf{0.5955}          & 0.6487 & 0.6223  & 0.6109          & 0.6160    & \textbf{0.5884}          \\ \midrule
            \multicolumn{3}{c|}{Avg. Rank}                                                                & 4.7500          & 5.0833 & 4.9167  & \textbf{3.9167}          & 5.9167 & 6.4167  & 5.5000          & 5.7500    & \textbf{2.5833}          \\ \bottomrule
        \end{tabular}
    }
    \caption{The evaluation result in the context-guided time series forecasting task. The lower values for all metrics represent the better performance. The best results among dataset-specific and dataset-shared methods are highlighted in bold, respectively.}
    \label{tab05}
\end{table*}

The baselines are broadly categorized into two groups. The first group consists of models trained and predicted on a single dataset with fixed history and prediction lengths, including DLinear \cite{DLinear}, iTransformer \cite{iTransformer}, GPT4TS \cite{GPT4TS}, and TimeLLM \cite{TimeLLM}. GPT4TS and TimeLLM both utilize pre-trained LLMs as their backbone. The second group comprises foundational models capable of zero-shot forecasting, such as TimeGPT \cite{TimeGPT}, Moirai \cite{Moirai}, TimesFM \cite{TimesFM}, and Chronos \cite{Chronos}. For the foundational models available in different sizes, we use their most powerful versions. All baselines are evaluated based on our runs using the same hardware as ChatTime, except for the closed-source model TimeGPT, which requires official API calls. We use official implementations from GitHub and follow the hyperparameter configurations recommended in their papers. The prompt templates for ChatTime are provided in Appendix \ref{subsection:Zero-Shot-Time-Series-Forecasting-Prompt}.

The experimental results are summarized in Table \ref{tab04}. To avoid a few datasets dominating the results, we primarily compare the average MAE (the lower, the better) and the average Rank (the smaller, the better) across eight datasets. By fine-tuning an existing pre-trained LLM instead of training it from scratch, ChatTime achieves 99.9\% of the zero-shot prediction accuracy of the previous SOTA method, Chronos, using only 4\% of the data. Compared to the full-shot forecasting model, ChatTime also attains 90.9\% of the prediction accuracy of the previous SOTA method, GPT4TS. Although introducing LLMs brings some performance gains for GPT4TS and TimeLLM, they do not significantly outperform the simple linear model DLinear. This validates that current unimodal methods may be approaching their saturation point. To visually compare the differences between these baselines, we provide a showcase in Appendix \ref{subsection:Zero-Shot-Time-Series-Forecasting-Showcase}, where ChatTime continues to demonstrate its superiority.

\subsection{Context-Guided Time Series Forecasting}
\label{subsection:Context-Guided-Time-Series-Forecasting-Experiment}

For the multimodal context-guided time series forecasting task, we conduct experiments on the three datasets collected in Section \ref{subsection:Instruction-Fine-Tuning}. We segment each dataset adhering to the protocols outlined in Section \ref{subsection:Zero-Shot-Time-Series-Forecasting-Experiment}. The settings for history length, prediction length, and evaluation metric also remain consistent with Section \ref{subsection:Zero-Shot-Time-Series-Forecasting-Experiment}. Notably, due to the limited multimodal datasets, the instruction fine-tuning phase of ChatTime is performed on partial training sets of these three datasets. Although this deviates from the zero-shot setup, ChatTime still does not require separate training for different scenarios but utilizes shared model weights.

The baselines are similar to Section \ref{subsection:Zero-Shot-Time-Series-Forecasting-Experiment}, except for including TGForecaster \cite{TGForecaster}, which can handle textual information. Moreover, to verify the auxiliary role of context in time series forecasting, we specifically establish a comparison baseline, ChatTime-, which excludes textual input during forecasting. The prompt templates for ChatTime are provided in Appendix \ref{subsection:Context-Guided-Time-Series-Forecasting-Prompt}.

The experimental results are summarized in Table \ref{tab05}. With the incorporation of textual information, TGForecaster and ChatTime exhibit superior performance compared to other baselines. Owing to the synergistic integration of the two modalities, ChatTime even surpasses TGForecaster, which is trained independently on each dataset. Moreover, ChatTime significantly outperforms ChatTime- using only unimodal values, affirming the effectiveness of contextual assistance. We provide showcases in Appendix \ref{subsection:Context-Guided-Time-Series-Forecasting-Showcase} that further validate the substantial potential and utility of ChatTime.

\subsection{Time Series Question Answering}
\label{subsection:Time-Series-Question-Answering-Experiment}

\begin{table}[t]
    \centering
    \setlength{\tabcolsep}{1.5mm}
    {\small
       \begin{tabular}{@{}cc|ccccc@{}}
            \toprule
            \multicolumn{1}{c|}{Feat}                                         & Len    & GPT4   & GPT3.5  & GLM4   & LLaMA3  & ChatTime        \\ \midrule
            \multicolumn{1}{c|}{\multirow{4}{*}{\rotatebox{90}{Trend}}}       & 64     & 0.6532 & 0.3507  & 0.7319 & 0.6799  & \textbf{0.9011} \\
            \multicolumn{1}{c|}{}                                             & 128    & 0.7015 & 0.5846  & 0.7574 & 0.5855  & \textbf{0.9068} \\
            \multicolumn{1}{c|}{}                                             & 256    & 0.7482 & 0.5028  & 0.6377 & 0.6143  & \textbf{0.8843} \\
            \multicolumn{1}{c|}{}                                             & 512    & 0.6346 & 0.5903  & 0.6697 & 0.6753  & \textbf{0.8234} \\ \midrule
            \multicolumn{1}{c|}{\multirow{4}{*}{\rotatebox{90}{Volatility}}}  & 64     & 0.5585 & 0.5633  & 0.6797 & 0.6373  & \textbf{0.7874} \\
            \multicolumn{1}{c|}{}                                             & 128    & 0.4979 & 0.3839  & 0.4770 & 0.4756  & \textbf{0.6954} \\
            \multicolumn{1}{c|}{}                                             & 256    & 0.4624 & 0.4894  & 0.5418 & 0.5246  & \textbf{0.6228} \\
            \multicolumn{1}{c|}{}                                             & 512    & 0.3169 & 0.3796  & 0.4549 & 0.5261  & \textbf{0.5736} \\ \midrule
            \multicolumn{1}{c|}{\multirow{4}{*}{\rotatebox{90}{Season}}}      & 64     & 0.3518 & 0.3428  & 0.3366 & 0.3484  & \textbf{0.6639} \\
            \multicolumn{1}{c|}{}                                             & 128    & 0.3515 & 0.3952  & 0.3464 & 0.3958  & \textbf{0.6517} \\
            \multicolumn{1}{c|}{}                                             & 256    & 0.5283 & 0.5089  & 0.3892 & 0.4120  & \textbf{0.6463} \\
            \multicolumn{1}{c|}{}                                             & 512    & 0.4457 & 0.4889  & 0.3892 & 0.4127  & \textbf{0.6244} \\ \midrule
            \multicolumn{1}{c|}{\multirow{4}{*}{\rotatebox{90}{Outlier}}}     & 64     & 0.7230 & 0.4325  & 0.5359 & 0.7051  & \textbf{0.8773} \\
            \multicolumn{1}{c|}{}                                             & 128    & 0.6327 & 0.5940  & 0.5298 & 0.5694  & \textbf{0.9032} \\
            \multicolumn{1}{c|}{}                                             & 256    & 0.6795 & 0.4579  & 0.5019 & 0.5073  & \textbf{0.8593} \\
            \multicolumn{1}{c|}{}                                             & 512    & 0.6219 & 0.4996  & 0.2822 & 0.4085  & \textbf{0.7478} \\ \midrule
            \multicolumn{2}{c|}{Avg. Acc}                                              & 0.5567 & 0.4728  & 0.5163 & 0.5299  & \textbf{0.7605} \\ \midrule
            \multicolumn{2}{c|}{Avg. Rank}                                             & 3.0625 & 4.0625  & 3.6250 & 3.2500  & \textbf{1.0000} \\ \bottomrule
        \end{tabular}
    }
    \caption{The evaluation result in the time series question answering task. Higher values mean better performance for all metrics, except Rank, which is better when lower. The best results are highlighted in bold.}
    \label{tab06}
\end{table}

For the multimodal time series question answering task, we conduct experiments on the dataset synthesized in Section \ref{subsection:Instruction-Fine-Tuning}. We exclude the 25K samples utilized for the instruction fine-tuning of ChatTime and use the remaining data as the test set. Baselines for comparison are powerful generic pre-trained LLMs: GPT4 \cite{GPT4}, GPT3.5 \cite{GPT3}, GLM4 \cite{GLM4}, and LLaMA3-70B \cite{LLaMA3}. For the input formats of the time series, we employ two prompts suggested by LLMTIME \cite{LLMTIME}, as described in Section \ref{subsection:Model-Architecture}. For GLM4, we have tested both prompts and select the format like LLaMA, which yields better results. Except for LLaMA3, which uses API from Alibaba \cite{Alibaba}, the remaining baselines all use their official API. The prompt templates for ChatTime are provided in Appendix \ref{subsection:Time-Series-Question-Answering-Prompt}. Given the nature of feature recognition, we report the accuracy (Acc) as evaluation metric, with higher scores indicating better performance.

The experimental results are summarised in Table \ref{tab06}. To avoid a few datasets dominating the results, we primarily compare the average Acc (the higher, the better) and the average Rank (the smaller, the better) across four features. Although generic LLMs have shown impressive performance across various text tasks, their efficacy in time series comprehension remains suboptimal. ChatTime, not only preserves the inference capabilities of LLMs but also demonstrates a superior understanding of time series features. We also provide showcases in Appendix \ref{subsection:Time-Series-Question-Answering-Showcase}.

\subsection{Ablation Study}
\label{subsection:Ablation-Study}

\begin{figure}[t]
    \centering
    \resizebox{0.98\linewidth}{!}
    {
        \includegraphics{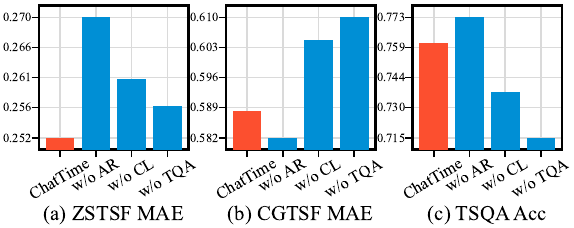}
    }
    \caption{The evaluation result between ChatTime and variants. Lower values are better for ZSTSF and CGTSF, while higher values are better for TSQA.}
    \label{fig02}
\end{figure}

To validate the soundness of each design in ChatTime, we perform an ablation study on the aforementioned three tasks and report their average results across all datasets individually. As depicted in Figure \ref{fig02}, we assess the indispensability of autoregressive continuous pre-training (w/o AR), clustering for time-series slices (w/o CL), and the text question answering in fine-tuning instructions (w/o TQA).

\begin{figure}[t]
    \centering
    \resizebox{0.98\linewidth}{!}
    {
        \includegraphics{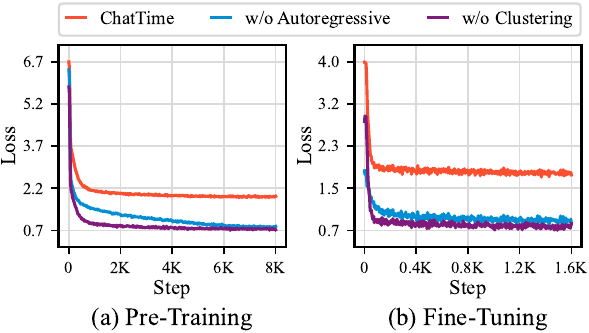}
    }
    \caption{The loss between ChatTime and its variants during continuous pre-training and instruction fine-tuning.}
    \label{fig03}
\end{figure}

In w/o AR, we substitute the 1M continuous pre-training dataset with the 100K instruction fine-tuning dataset. We increase the epoch by ten times to maintain consistent parameter iterations. This increase leads to a slight improvement in CGTSF and TSQA. However, removing the pre-training dataset causes a struggle to grasp the fundamental time series features, significantly reducing the zero-shot inference capability and practical value. The loss depicted in Figure \ref{fig03} also illustrates the overfitting in ChatTime after replacing time series pre-training data.

In the w/o CL, we substitute the high-quality time series slices obtained from clustering with low-quality data randomly sampled from the 10M original slices. The findings indicate that ChatTime lacks sufficient comprehension of time series after replacement. There are various degradations across the three tasks. The loss observed in Figure \ref{fig03} also confirms that randomly sampled data is less challenging to model, making ChatTime prone to overfitting.

In the w/o TQA, we exclude the text question answering dataset in the instruction fine-tuning phase. The findings indicate that omitting this task hampers the inference capability, resulting in performance degradation across all three tasks, particularly in the multimodal CGTSF and TSQA.

\section{Conclusion}
\label{section:Conclusion}

In this study, we concentrate on the efficient construction of a multimodal time series foundation model that allows for zero-shot inference and supports both time series and textual bimodal inputs and outputs. By innovatively characterizing time series as a foreign language, we introduce ChatTime, a framework for the unified processing of time series and text. To validate the superior performance of ChatTime, we have meticulously designed a series of experiments and constructed four multimodal datasets to fill relevant data gaps. The experimental results demonstrate the significant potential and utility of ChatTime, offering novel perspectives and solutions for time series analysis tasks. Due to resource constraints, ChatTime has not yet reached saturation. In future work, we plan to use more data and computational resources to further extend its applicable tasks, such as anomaly detection, classification, or summarization.

\section*{Acknowledgments}
\label{section:Acknowledgments}

This work was supported in part by the National Natural Science Foundation of China under Grants (62171057, 62201072, 62471055, U23B2001, 62321001, 62101064), the Ministry of Education and China Mobile Joint Fund (MCM20200202, MCM20180101), the Fundamental Research Funds for the Central Universities (2024PTB-004), and the BUPT Excellent Ph.D. Students Foundation (CX20241016).

%%%%%%%%%%%%%%%%%%%%%%%%%%%%%%%%%%%%%%%%%%%%%%%%%%%%%%%%%%%%

\bibliography{aaai25}

%%%%%%%%%%%%%%%%%%%%%%%%%%%%%%%%%%%%%%%%%%%%%%%%%%%%%%%%%%%%

\clearpage
\appendix

\section{Prompt}
\label{section:Prompt}

\subsection{Zero-Shot Time Series Forecasting}
\label{subsection:Zero-Shot-Time-Series-Forecasting-Prompt}

\begin{figure}[h]
    \centering
    \resizebox{0.95\linewidth}{!}
    {
        \includegraphics{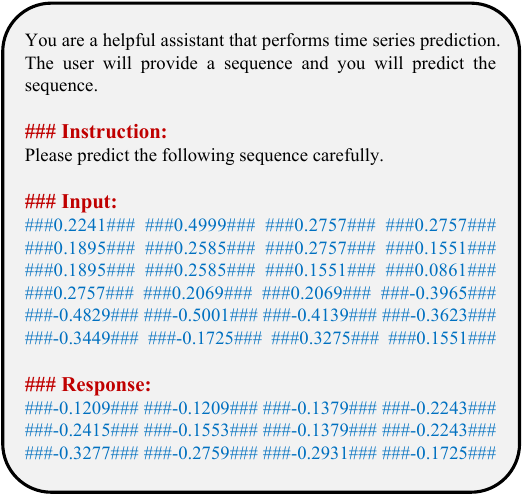}
    }
    \caption{The prompt in the conventional unimodal time series forecasting.}
    \label{fig04}
\end{figure}

As illustrated in Figure \ref{fig04}, we meticulously design the prompts for the zero-shot time series forecasting task. The comprehensive prompt comprises a system prompt, an introduction, an input, and a response. The system prompt establishes the role of ChatTime and provides a general task description. The introduction delineates the specific task details. The input and the response represent the history and prediction series, translated into foreign language words. The prediction series are supplied only during the instruction fine-tuning phase and are concealed during inference to stimulate ChatTime generation. The system prompt and introduction remain consistent throughout the zero-shot time series forecasting task. The variable components are the history and prediction series in the input and response.

\subsection{Context-Guided Time Series Forecasting}
\label{subsection:Context-Guided-Time-Series-Forecasting-Prompt}

\begin{figure}[t]
    \centering
    \resizebox{0.93\linewidth}{!}
    {
        \includegraphics{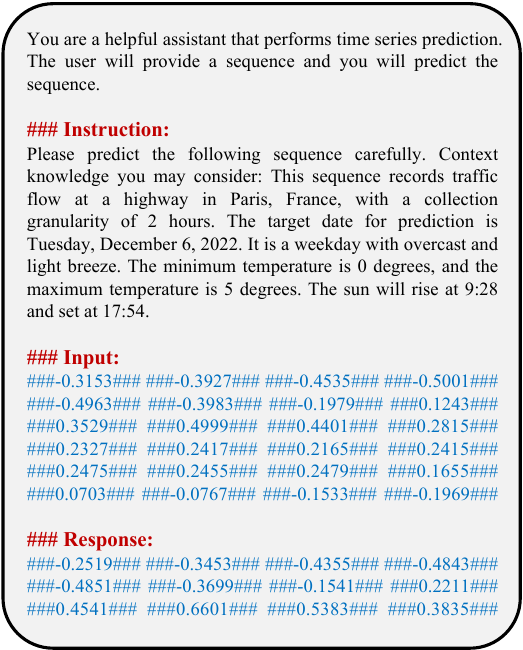}
    }
    \caption{The prompt in the context-guided time series forecasting task.}
    \label{fig05}
\end{figure}

As illustrated in Figure \ref{fig05}, we meticulously design the prompts for the context-guided time series forecasting task. The comprehensive prompt comprises a system prompt, an introduction, an input, and a response. The system prompt establishes the role of ChatTime and provides a general task description. The introduction outlines the specific task details and offers contextual knowledge with supplementary information. The input and the response represent the history and prediction series, translated into foreign language words. The prediction series are supplied only during the instruction fine-tuning phase and are concealed during inference to stimulate ChatTime generation. The system prompt and task description in the introduction remain consistent throughout the context-guided time series forecasting task. The variable components are the contextual knowledge in the introduction, as well as the history and prediction series in the input and response.

\begin{figure}[tp]
    \centering
    \resizebox{0.935\linewidth}{!}
    {
        \includegraphics{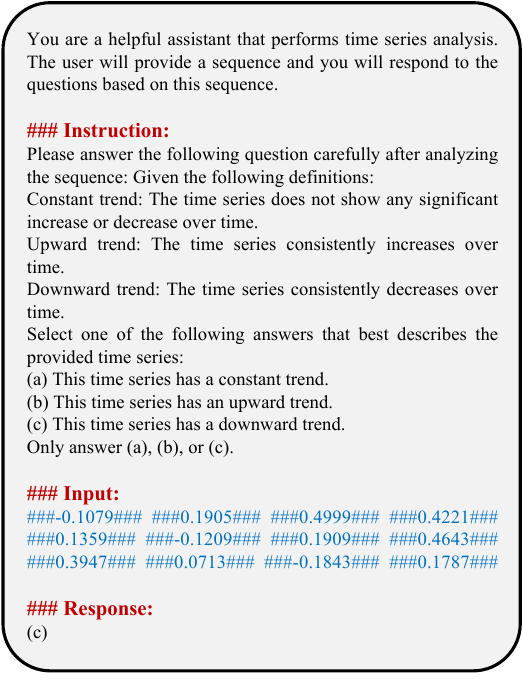}
    }
    \caption{The prompt of trend feature in the time series question answering task.}
    \label{fig06}
\end{figure}

\begin{figure}[bp]
    \centering
    \resizebox{0.94\linewidth}{!}
    {
        \includegraphics{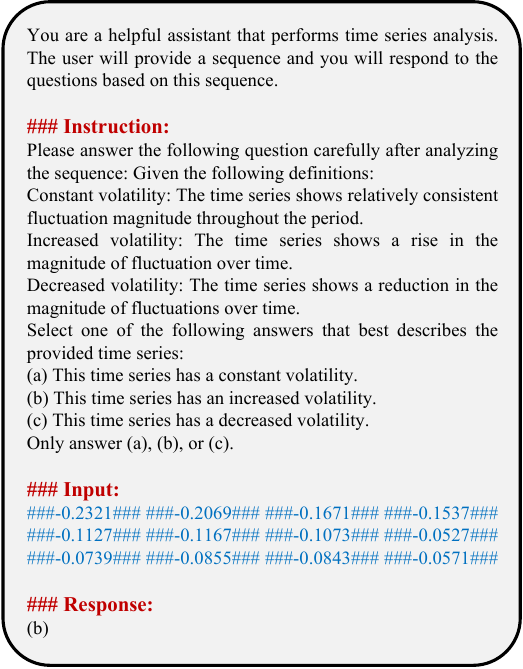}
    }
    \caption{The prompt of volatility feature in the time series question answering task.}
    \label{fig07}
\end{figure}

\begin{figure}[tp]
    \centering
    \resizebox{0.942\linewidth}{!}
    {
        \includegraphics{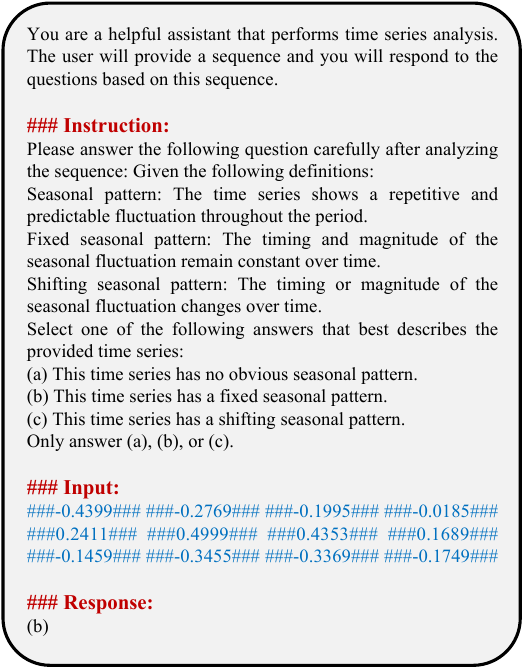}
    }
    \caption{The prompt of season feature in the time series question answering task.}
    \label{fig08}
\end{figure}

\begin{figure}[bp]
    \centering
    \resizebox{0.91\linewidth}{!}
    {
        \includegraphics{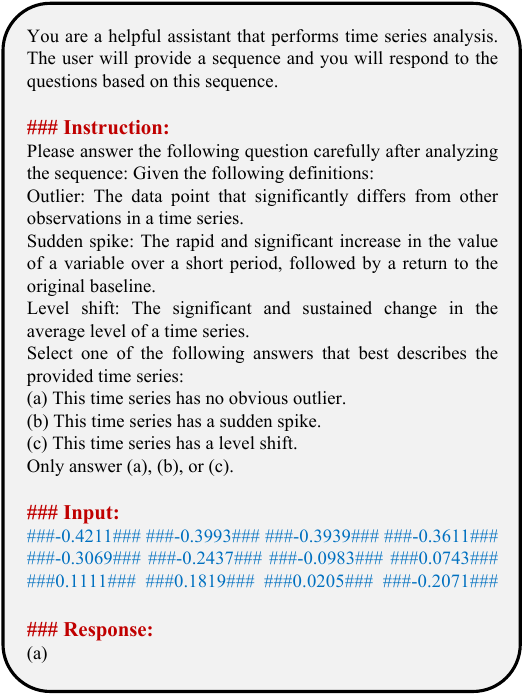}
    }
    \caption{The prompt of outlier feature in the time series question answering task.}
    \label{fig09}
\end{figure}

\subsection{Time Series Question Answering}
\label{subsection:Time-Series-Question-Answering-Prompt}

As illustrated in Figures \ref{fig06}, \ref{fig07}, \ref{fig08}, and \ref{fig09}, we meticulously design the prompts for the time series question answering task. The comprehensive prompt comprises a system prompt, an introduction, an input, and a response. The system prompt establishes the role of ChatTime and provides a general task description. The introduction outlines the specific task details and provides additional background knowledge to help ChatTime understand the typical features of a time series. The input provides the time series to be analyzed after being translated into foreign language words. The response part is the correct answer. The correct answer is supplied only during the instruction fine-tuning phase and is concealed during inference to stimulate ChatTime generation. The system prompt and task description in the introduction remain consistent throughout the time series question answering task. The variable components are the background knowledge in the introduction, the time series to be analyzed in the input, and the correct answers in the response.

\section{Dataset}

\subsection{Zero-Shot Time Series Forecasting}
\label{subsection:Zero-Shot-Time-Series-Forecasting-Dataset}

\begin{table}[t]
    \centering
    \setlength{\tabcolsep}{2mm}
    {\small
        \begin{tabular}{@{}c|cccc@{}}
            \toprule
            Dataset        & Length & Frequency  & Information    \\ \midrule
            Electric       & 26304  & 1 Hour     & Energy         \\
            Exchange       & 7588   & 1 Day      & Finance        \\
            Traffic        & 17544  & 1 Hour     & Transportation \\
            Weather        & 52696  & 10 Minutes & Climate        \\
            ETTh1 \& ETTh2 & 17420  & 1 Hour     & Energy         \\
            ETTm1 \& ETTm2 & 69680  & 15 Minutes & Energy         \\ \bottomrule
        \end{tabular}
    }
    \caption{The statistics of each dataset in the traditional unimodal time series forecasting task.}
    \label{tab07}
\end{table}

\begin{table}[t]
    \centering
    \setlength{\tabcolsep}{4mm}
    {\small
        \begin{tabular}{@{}c|cccc@{}}
            \toprule
            Dataset     & Length & Frequency  & Information    \\ \midrule
            MSPG        & 38016  & 15 Minutes & Energy         \\
            LEU         & 35088  & 30 Minutes & Energy         \\
            PTF         & 8760   & 1 Hour     & Transportation \\ \bottomrule
        \end{tabular}
    }
    \caption{The statistics of each dataset in the context-guided time series forecasting task.}
    \label{tab08}
\end{table}

For the regular unimodal time series forecasting task, we conduct extensive experiments on eight real-world datasets across four domains: Electric, Exchange, Traffic, and Weather, in addition to four ETT datasets. Table \ref{tab07} summarizes the statistics of these datasets. These datasets have been widely utilized for benchmarking purposes and are publicly available. (1) Electric comprises hourly electricity consumption for 321 customers from 2012 to 2014. (2) Exchange encompasses panel data on daily exchange rates for 8 countries from 1990 to 2019. (3) Traffic aggregates hourly road occupancy rates measured by 862 sensors on San Francisco Bay Area freeways from 2015 to 2016. (4) Weather captures 21 weather parameters monitored every 10 minutes from Germany in 2020. (5) ETT records the oil temperature and load characteristics of two power transformers from 2016 to 2018, each at 2 different resolutions, resulting in a total of four datasets: ETTm1, ETTm2, ETTh1, and ETTh2.

\subsection{Context-Guided Time Series Forecasting}
\label{subsection:Context-Guided-Time-Series-Forecasting-Dataset}

The context-guided time series forecasting task entails the transformation of text into time series data. Relevant multimodal datasets are limited. To address these data gaps, we have collected three multimodal datasets that offer valuable resources for future research. Table \ref{tab08} summarizes the statistics of these datasets. MSPG comprises 13 months of solar power generation data on 27 photovoltaic sites in Melbourne from 2021 to 2022. LEU encompasses 24 months of electricity usage data on 16 households in London from 2012 to 2013. PTF includes 12 months of traffic flow data on 32 traffic detectors in Paris during 2012. We gather raw time series records from Kaggle, a prominent open-source platform. To prevent future data leakage, we incorporate only background, weather, and date as textual auxiliary information. The background includes a description of the dataset and its collection granularity. Weather encompasses forecast data obtained from Open-Meteo, including weather codes, temperatures, and sunrise and sunset times. Regarding dates, we include the raw date, day of the week, and holiday information. All auxiliary data is concatenated into coherent text and strictly aligned with the time series records by day.

\subsection{Time Series Question Answering}
\label{subsection:Time-Series-Question-Answering-Dataset}

\begin{table}[t]
    \centering
    \setlength{\tabcolsep}{3mm}
    {\small
        \begin{tabular}{@{}c|cccc@{}}
            \toprule
            Feature       & Category  & Number & Length\\ \midrule
            Trend         & 3         & 12000  & \{64, 128, 256, 512\} \\
            Volatility    & 3         & 12000  & \{64, 128, 256, 512\} \\
            Season        & 3         & 12000  & \{64, 128, 256, 512\} \\
            Outlier       & 3         & 12000  & \{64, 128, 256, 512\} \\ \bottomrule
        \end{tabular}
    }
    \caption{The statistics of each feature in the time series question answering dataset.}
    \label{tab09}
\end{table}

In the time series question answering task, we formulate question and answer pairs based on identifying four generic typical time series features, which aid ChatTime in comprehending the fundamental principles of time series. Table \ref{tab09} summarizes the statistics of this dataset. Trend encompasses three categories: upward trend, downward trend, and constant trend. Volatility includes three categories: increased volatility, decreased volatility, and constant volatility. Season is categorized into three groups: fixed seasonality, shifting seasonality, and no seasonality. Outliers feature three categories: sudden spike, level shift, and stable no outlier. We use KernelSynth to generate time series slices of four lengths, \{64, 128, 256, 512\}, to enhance robustness. By aligning time series features with textual representations, this task can aslo improve the performance of ChatTime in various time series downstream tasks.

\section{Showcase}
\label{section:Showcase}

\begin{figure*}[tp]
    \centering
    \resizebox{0.97\linewidth}{!}
    {
        \includegraphics{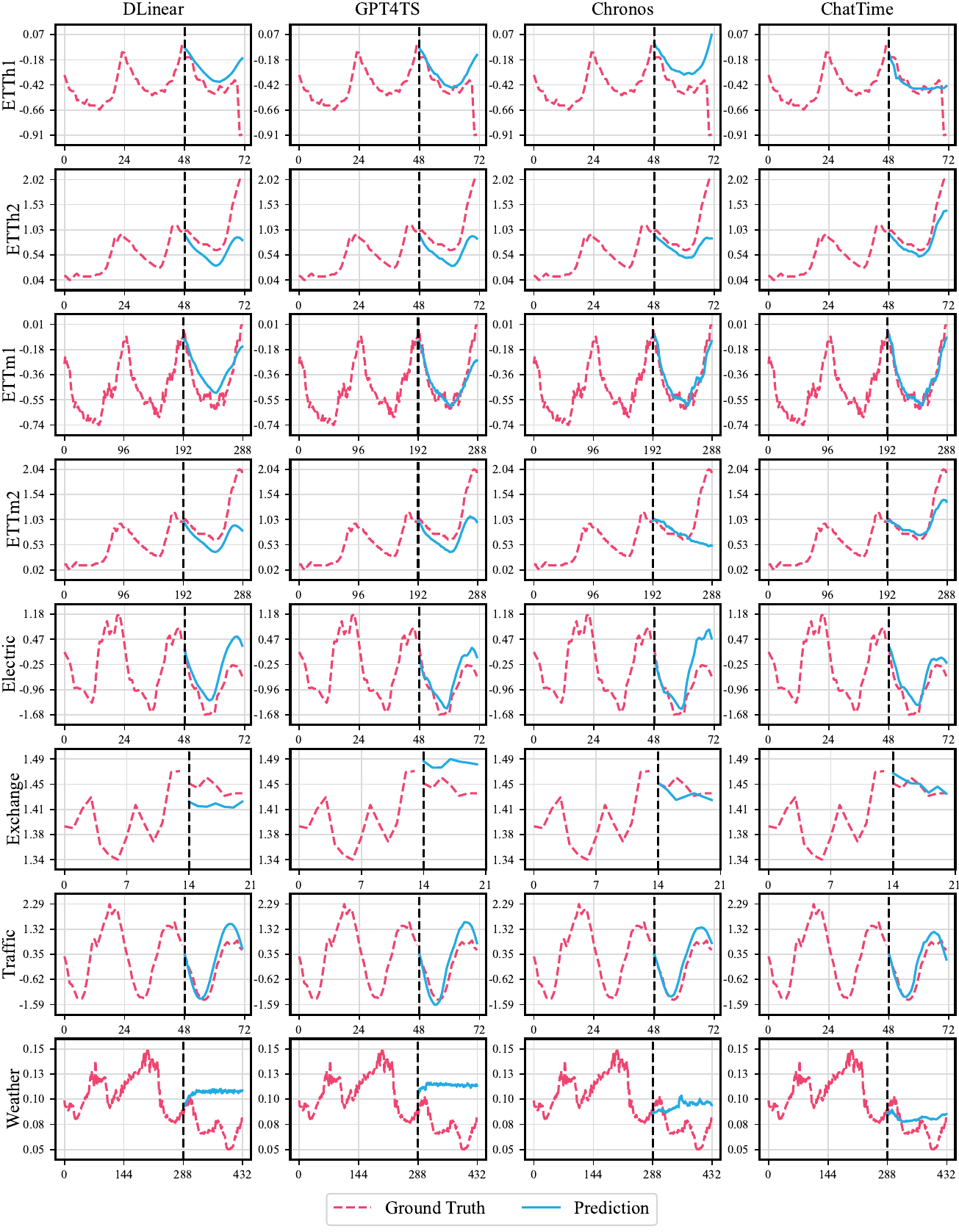}
    }
    \caption{The showcase in the traditional unimodal time series forecasting task.}
    \label{fig10}
\end{figure*}

\begin{figure*}[tp]
    \centering
    \resizebox{\linewidth}{!}
    {
        \includegraphics{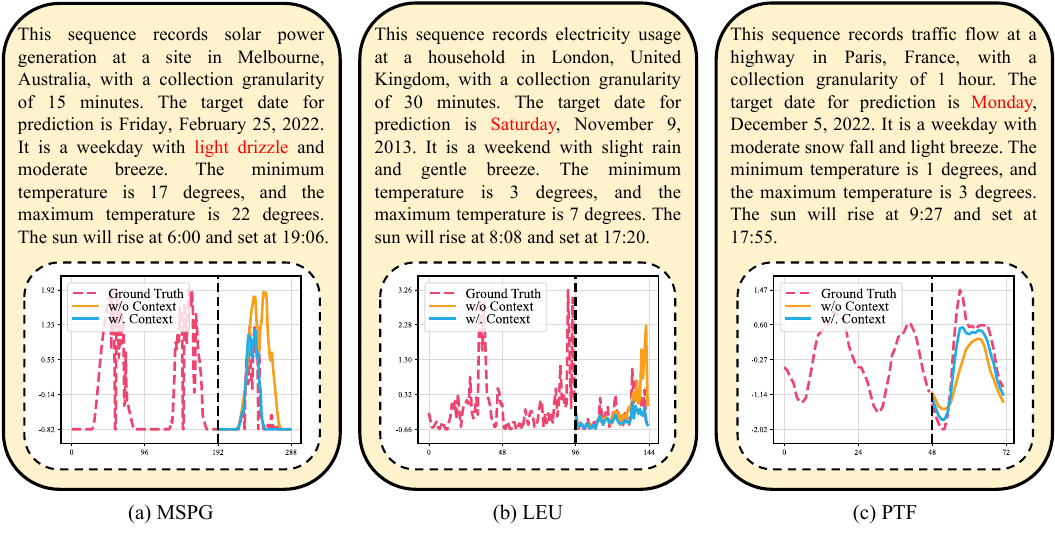}
    }
    \caption{The showcase in the context-guided time series forecasting task.}
    \label{fig11}
\end{figure*}

\begin{figure*}[bp]
    \centering
    \resizebox{\linewidth}{!}
    {
        \includegraphics{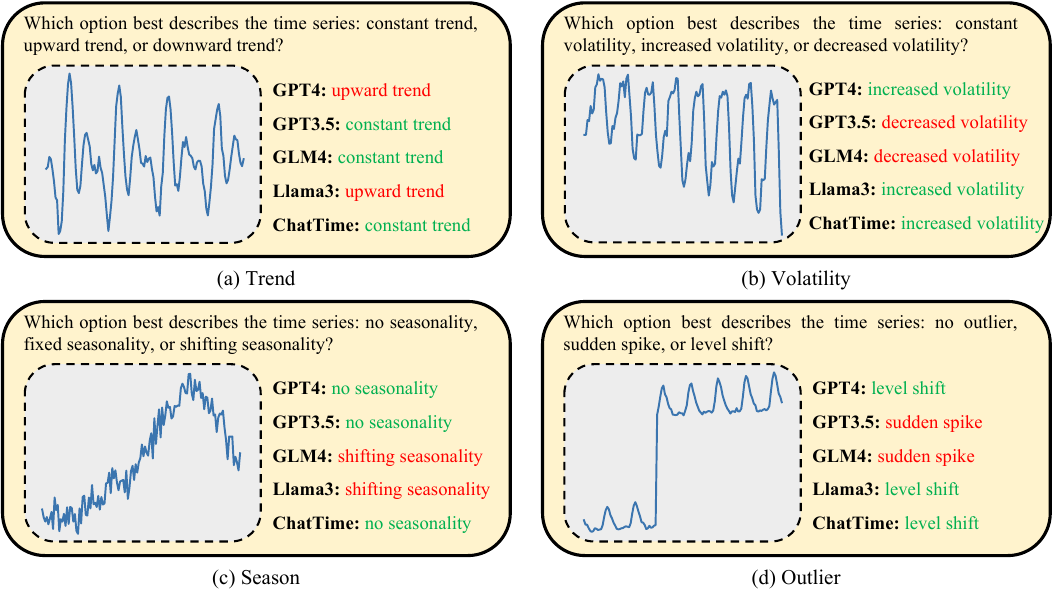}
    }
    \caption{The showcase in the time series question answering task.}
    \label{fig12}
\end{figure*}

\subsection{Zero-Shot Time Series Forecasting}
\label{subsection:Zero-Shot-Time-Series-Forecasting-Showcase}

In addition to evaluation metrics, forecasting quality is crucial. To provide a clear comparison between ChatTime and SOTA forecasting baselines, we present showcases for eight real-world benchmark datasets in Figure \ref{fig10}. The length of the history window for all datasets is set to twice the length of the prediction window. In the full-shot forecasting models, the prediction performance of the simple linear model DLinear is comparable to that of the current SOTA complex model GPT4TS. This indicates that current unimodal approaches may be reaching a saturation point. In the zero-shot forecasting models, our proposed ChatTime uses only 4\% of the pre-training data to achieve accuracy similar to that of the current SOTA foundation Chronos, and even surpasses it in some scenarios. Based on extensive pre-training on time series data, the foundation models have a deeper understanding of the fundamental principles of time series. In non-stationary scenarios, ChatTime often provides more accurate forecasting trends than models trained on a single dataset, such as ETTh2 and Electric, validating the significant potential and utility of the generic foundation model.

\subsection{Context-Guided Time Series Forecasting}
\label{subsection:Context-Guided-Time-Series-Forecasting-Showcase}

Recent studies have demonstrated that the prediction performance of simple linear models can often rival that of SOTA complex models, indicating that current unimodal approaches may be nearing their performance limits. To achieve higher prediction accuracy, models must incorporate additional auxiliary information. ChatTime is fine-tuned by expanding the vocabulary based on the pre-trained LLM LLaMA. It supports seamless input of both time series and textual bimodality while retaining the powerful inference capability of LLaMA. To validate the superiority of ChatTime in context-guided forecasting tasks, we present prediction cases on three real-world benchmark datasets in Figure \ref{fig11}. In the MSPG scenario, rain accompanied by cloud cover reduces solar power generation. In the LEU scenario, a weekend break leads to lower electricity consumption. In the PTF scenario, weekday travel results in higher traffic flow. Without context-guided information, ChatTime relies solely on unimodal information, such as history time series, to make limited predictions. It cannot perceive the impact of various external events. However, after providing context-guided information, the prediction accuracy of ChatTime on all three datasets is significantly improved. These cases further confirm the necessity of context-guided forecasting.

\subsection{Time Series Question Answering}
\label{subsection:Time-Series-Question-Answering-Showcase}

Grasping the fundamental principles of time series is essential for executing downstream tasks. To validate the superiority of ChatTime in time series comprehension, we present question answering examples for four typical time series features in Figure \ref{fig12}. Notable differences exist between time series and text. While generic pre-trained LLMs have achieved remarkable success in various textual tasks, their performance in time series comprehension is less satisfactory. ChatTime, fine-tuned by expanding the vocabulary based on the pre-trained LLM LLaMA, not only provides an excellent understanding of time series features but also supports seamless bimodal input and output of both time series and text. This significantly broadens its task range, enabling both time series question answering and summarization.

\end{document}